\documentclass[journal]{new-aiaa}

\usepackage{graphicx}
\usepackage{amsmath}
\usepackage{float}
\usepackage{booktabs}  % Defines \toprule, \midrule, \bottomrule
\usepackage{adjustbox} % Defines the adjustbox environment
\usepackage{textcomp}  % Provides \textmu
\usepackage{xcolor}    % Required for color commands
\usepackage{pifont}
\usepackage{capt-of} % Allows table captions without floating
\usepackage{array}
\usepackage{makecell}
\usepackage{stfloats}
\fnbelowfloat

\newcommand{\cmark}{\textcolor{green!70!black}{\ding{51}}}
\newcommand{\xmark}{\textcolor{red}{\ding{55}}}

\newcolumntype{V}{!{\vrule width 1.2pt}}

\newcommand{\affilECE}{\textsuperscript{*}}
\newcommand{\affilAV}{\textsuperscript{\textdagger}}
\newcommand{\affilME}{\textsuperscript{\textdaggerdbl}}

\begin{document}

\author{
Derek Ward\affilECE,
Brian Bascom\affilECE,
Tyler Larsen\affilECE,
James Usevitch\affilECE,
Phil Tokumaru\affilAV,
and Tim McLain\affilME
}

\title{Veloxity: Towards Memory-Safe Embedded Autopilot Firmware using the Rust Programming Language\vspace{-15pt}}

%\author[1]{Derek Ward\footnote{
%Electrical and Computer Engineering Department,
%Brigham Young University, Provo, UT, USA;
%wderekb@byu.edu.}}
%
%\author[1]{Brian Bascom\footnote{
%Electrical and Computer Engineering Department,
%Brigham Young University, Provo, UT, USA;
%bascomb1@byu.edu.}}
%
%\author[1]{Tyler Larsen\footnote{
%Electrical and Computer Engineering Department,
%Brigham Young University, Provo, UT, USA;
%tylerrl2@byu.edu.}}
%
%\author[1]{James Usevitch\footnote{
%Electrical and Computer Engineering Department,
%Brigham Young University, Provo, UT, USA;
%james\_usevitch@byu.edu.}}
%
%\author[2]{Phil Tokumaru\footnote{
%%AeroVironment Inc., Moorpark, CA, USA;
%tokumaru@avinc.com.}}
%
%\author[1]{Tim McLain\footnote{
%Mechanical Engineering Department,
%Brigham Young University, Provo, UT, USA;
%mclain@byu.edu.}}
%
%
%
%\affil[1]{Brigham Young University, Provo, UT, USA}
%\affil[2]{AeroVironment Inc., Moorpark, CA, USA}

\maketitle

% Affiliations and emails at the bottom of page 1
\begingroup
\renewcommand{\thefootnote}{}
\makeatletter
\renewcommand{\@makefntext}[1]{\noindent#1}
\makeatother

\footnotetext[0]{%
  \singlespacing
  \fontsize{9}{10.5}\selectfont
  \raggedright
  \setlength{\tabcolsep}{0pt}
  \renewcommand{\arraystretch}{1.0}

  \begin{tabular}{@{}r@{\hspace{0.6em}}p{\dimexpr\linewidth-2em\relax}@{}}
    \affilECE &
    Department of Electrical and Computer Engineering,
    Brigham Young University, Provo, UT 84602, USA \\
    & \textit{\{wderekb, bascomb1, tylerrl2, james\_usevitch\}@byu.edu} \\[2pt]

    \affilAV &
    AeroVironment Inc., Moorpark, CA 93021, USA \\
    & \textit{tokumaru@avinc.com} \\[2pt]

    \affilME &
    Department of Mechanical Engineering,
    Brigham Young University, Provo, UT 84602, USA \\
    & \textit{mclain@byu.edu}
  \end{tabular}
}
\endgroup

\vspace{-20pt}

\begin{figure}[htbp]
    \centering
    \makebox[\linewidth][c]{%
        \includegraphics[width=0.9\linewidth]{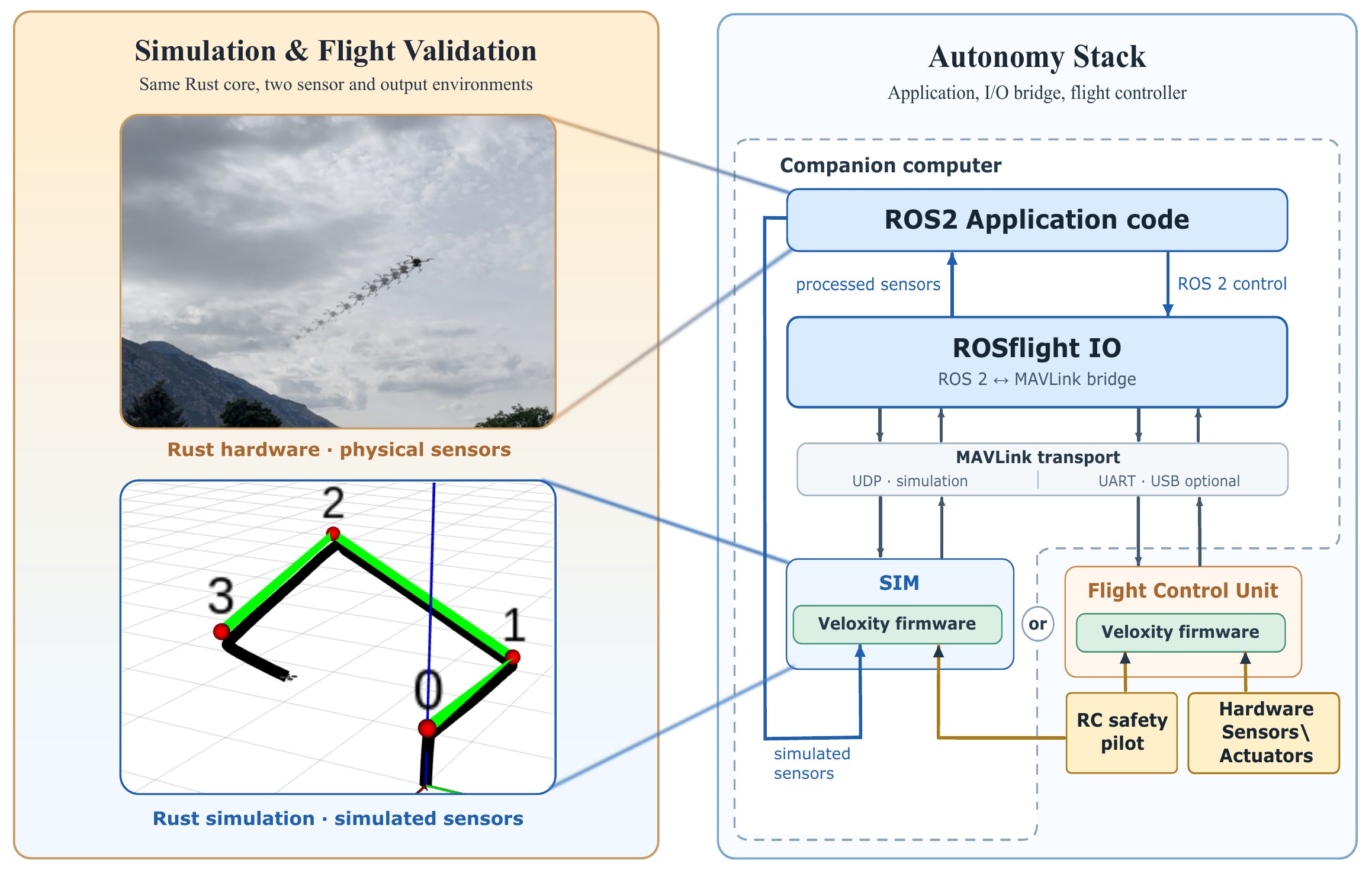}%
    }
    \caption{The Veloxity deployment architecture. Veloxity is a memory-safe, Rust-based flight control firmware library for unmanned aerial vehicles supporting simulated and hardware flight.
    }
    \label{fig:overview}
\end{figure}

% \vspace{-10pt}

\vspace{1em}

%%%%%%%%%%%%%%%%%%%%%%%%%%%%%%%%%%%%%%%%%%%%%%%%%%%%%%%%%%%%%%%%%%%%%%%%%%%%%%%%
\begin{abstract}

% Abstract can be 200 words max (https://aiaa.org/publications/journals/Journal-Author#follow-the-minimum-formatting-requirements)

\textsc{Abstract:} 
Recent years have witnessed increased momentum towards memory-safe programming languages for safety-critical systems. Memory-safe languages can eliminate entire classes of common software errors that have historically required extensive analysis and expenditure to avoid. Nevertheless, memory-safe language adoption in aerospace remains fragmented and incomplete, particularly for unmanned aerial vehicles (UAVs). Recent Rust UAV projects either run only in simulation, are closed source, or have not been integrated with an established autonomy ecosystem and tested against existing C/C++ firmware. To address this gap, we present Veloxity: an open-source embedded UAV autopilot written in the memory-safe Rust programming language. Veloxity lowers the barrier to using memory-safe languages within modern ecosystems by providing a minimal, modular foundation for autonomous UAVs. We compare Veloxity to equivalent autopilots written in C++ in simulation, with hardware testbenches, and via outdoor waypoint-following flight tests, demonstrating functional parity with improved performance. These results indicate that modern memory-safe languages are a viable option for reliable and performant firmware for autonomous aerospace systems.

\end{abstract}

\begingroup
\centering

% Shared vertical centering for all labels and headers
\newcommand{\centeredcell}[1]{%
    \raisebox{-0.1\height}{%
        \vbox to 12pt{%
            \vfil
            \hbox{#1}%
            \vfil
        }%
    }%
}

% First-column labels
\newcommand{\rowlabel}[2]{%
    \centeredcell{%
        \parbox[c]{1.65cm}{%
            \centering
            \scriptsize\bfseries
            \setlength{\baselineskip}{8pt}%
            #1\\[-1pt]#2%
        }%
    }%
}

% Two-line column headers
\newcommand{\colhead}[2]{%
    \centeredcell{%
        {\footnotesize\bfseries
        \shortstack[c]{#1\\[-1pt]#2}}%
    }%
}

% Single-line column headers
\newcommand{\singlehead}[1]{%
    \centeredcell{%
        {\footnotesize\bfseries #1}%
    }%
}

% \par\nobreak\vspace{5pt}

% Table dimensions
\renewcommand{\arraystretch}{1.0}
\setlength{\tabcolsep}{4pt}

\begin{table}[!b]
\centering
\noindent\resizebox{\textwidth}{!}{%
\begin{tabular}{|c|c|c|c|c|c|c|c|cV c|}
    \hline

    &
    \singlehead{PX4~\cite{px4repo}} &
    \singlehead{ArduPilot\cite{ardupilot}} &
    \colhead{Fusion}{Reflex\cite{fusionReflex}} &
    \singlehead{Crazyflie\cite{bitcraze2026crazyflie}} &
    \singlehead{Peng\cite{pengQuad}} &
    \singlehead{Adamant\cite{adamantFramework}} &
    \singlehead{Cybflight\cite{lin2026cybflight}} &
    \singlehead{ROSflight\cite{moore2025rosflight2}} &
    \singlehead{Veloxity}
    \\ \hline

    \rowlabel{Open}{Source} &
    \cmark & \cmark & \xmark & \cmark &
    \cmark & \cmark & \cmark & \cmark & \cmark
    \\ \hline

    \rowlabel{ROS~2}{Integration} &
    \cmark & \cmark & \xmark & \cmark &
    \xmark & \xmark & \xmark & \cmark & \cmark
    \\ \hline

    \rowlabel{Memory}{Safe} &
    \xmark & \xmark & \cmark & \xmark &
    \cmark & \cmark & \cmark & \xmark & \cmark
    \\ \hline

    \rowlabel{SIL}{Simulation} &
    \cmark & \cmark & \xmark & \cmark &
    \cmark & \xmark & \xmark & \cmark & \cmark
    \\ \hline

    \rowlabel{Hardware}{Validated} &
    \cmark & \cmark & \cmark & \cmark &
    \xmark & \cmark & \cmark & \cmark & \cmark
    \\ \hline

\end{tabular}%
}

% Caption
\captionof{table}{Comparison of UAV flight software frameworks. Memory safe: flight software written in a memory-safe language, apart from low-level hardware access. ROS~2 integration: a native interface to ROS~2 autonomy software. SIL Integration: proof of software-in-the-loop (SIL) simulation. Hardware validated: demonstrated on physical flight hardware.\vspace{-1pt}}
\label{tab:flight_frameworks}
\end{table}

\par
\endgroup

\section{Introduction}

\lettrine{A}{dvances} in autonomy research have expanded the safety assurances of unmanned aerial vehicles (UAVs), but the low-level software on which safe flight ultimately depends is still largely written in unsafe languages.
In small UAVs, timing-sensitive sensor processing, estimation, control, mixing, and actuator output remain on an embedded flight controller, making the robustness of the embedded firmware paramount. Defects in the firmware (particularly those related to program memory use) can result in loss of vehicle and damage to livelihood.
Most open-source flight-control firmware is implemented in C or C++ \cite{surveyOfUAVAutopilotsChao2010,surveyOfOpenSourceAutopilots2024}, languages that provide limited compile-time protection against memory errors and unsynchronized shared-state access.
Memory-safety violations such as out-of-bounds accesses, use-after-free errors, and data races can corrupt program state or terminate execution.
Because their effects may emerge far from the original defect and depend on a particular execution timing, these faults can be difficult to reproduce (and thus difficult to diagnose).
In flight firmware, the resulting corruption or loss of execution can affect sensing, state estimation, control, or actuator output while the vehicle is airborne.
Preventing such faults largely relies on development practices including coding standards, review, static analysis, and testing.

High-assurance languages and toolsets such as Ada and SPARK offer stronger verification, but obtaining that assurance can impose substantial development tradeoffs.
It can require developers to define proof obligations, work within language restrictions, and maintain specialized verification expertise.
These costs have not prevented the use of Ada and SPARK in high-assurance aerospace software or targeted flight-control demonstrations.
For example, Adamant provides a model-based, platform-agnostic Ada framework for embedded spacecraft flight software, including multitasking and support for SPARK proofs \cite{adamantFramework}. The Crazyflie's stock firmware is written in C \cite{bitcraze2026crazyflie}, but a separate demonstration rewrote the vehicle's stabilization subsystem in SPARK, proved it free of runtime errors, and flew it with the remaining C drivers and FreeRTOS infrastructure unchanged \cite{gracio2015crazyflieSpark}.
However, the proof effort, specialized expertise, and comparatively small development ecosystem associated with Ada and SPARK can impede widespread adoption in hobbyist and open-source UAV communities, where rapid iteration and integration with existing software are priorities.
This leaves room for approaches that balance compile-time protection with rapid development and compatibility with established research ecosystems.

Rust \cite{rustbook2026borrowing} is a modern programming language built to enforce memory access constraints through ownership rules and its type system, while retaining much of the familiar syntax of C/C++. In Rust, memory management can largely be checked and verified at compile time, but maintains execution speeds similar to C/C++ that are required for low-level control and resource-constrained embedded software.
These protections do not depend on a garbage collector, and Rust's \texttt{no\_std} environment supports systems without a conventional operating-system runtime.
Rust is also favorable for AI-assisted development due to the detailed compiler diagnostics capabilities that code-generation systems can use to perform iterative development and checking.
Perhaps the most compelling motivation Rust provides is the evidence from large companies that these safety benefits do not come with penalties in productivity or development.
For instance, in a survey of more than 1,000 Google developers who contributed Rust code (many of whom came from C++), Google found no evidence of lower productivity in Rust relative to the languages those developers had previously used; more than half reported becoming as productive in Rust within four months \cite{bergstromBrennan2023rustJourney}.
Despite the advantages, evidence remains limited for Rust in embedded flight firmware or for its evaluation within an established autonomy ecosystem. 

Rust has only recently begun to appear in UAV software. Peng \cite{pengQuad} is a Rust quadrotor framework for simulation, trajectory planning, and control, but does not run on flight hardware. Fusion Reflex \cite{fusionReflex} is a commercial flight controller whose Linux-hosted software is written in Rust, but it is not open source. Most recently, Cybflight \cite{lin2026cybflight} runs a modular Rust autopilot for calculating state estimation and nonlinear control directly on an STM32H743 flight controller. None of these, however, operate within an existing ROS~2 \cite{macenski2022ros2} autonomy ecosystem or present a comparison against an established C or C++ firmware. To our knowledge, no published literature has compared the use of Rust to more common alternatives in embedded UAV firmware. To overcome this shortcoming, we present the following contributions:

\begin{itemize}
\item An open-source UAV firmware autopilot implementation written in Rust. This includes a novel board- and protocol-independent \texttt{no\_std} core architecture supporting embedded and simulated runtimes, with a priority-based scheduler that prioritizes high-rate inertial measurement unit (IMU) processing and control.
\item An empirical comparison with the ROSflight firmware (C/C++) it was modeled after at multiple integration boundaries: flash footprint, communication latency, numerical agreement, and SIL waypoint following within the Robot Operating System 2 (ROS~2) ROScopter autonomy stack.
\item A real-time hardware implementation and field demonstration of Veloxity on STM32H743-based Pixracer Pro flight control unit (FCU) commonly used for flight-control research (Figs.~\ref{fig:overview} and~\ref{fig:outdoor_vehicle}).
\end{itemize}

% \section{Background and Design Motivation}
\section{ROSflight Architecture}

ROSflight \cite{jackson2020rosflight,moore2025rosflight2} is a lean, open-source autopilot ecosystem designed to reduce the effort required to move UAV research from simulation to hardware.
It was developed to reduce the barrier to entry for conducting highly customized autopilot research by favoring a modular, documented, and deliberately limited architecture over the broader feature sets of modern autopilots (e.g., PX4 and ArduPilot \cite{ardupilot,px4repo}). Veloxity extends the ROSflight ecosystem with memory safe firmware guarantees.

\begin{figure}[H]
    \centering
    \makebox[\linewidth][c]{%
        \includegraphics[width=\linewidth]{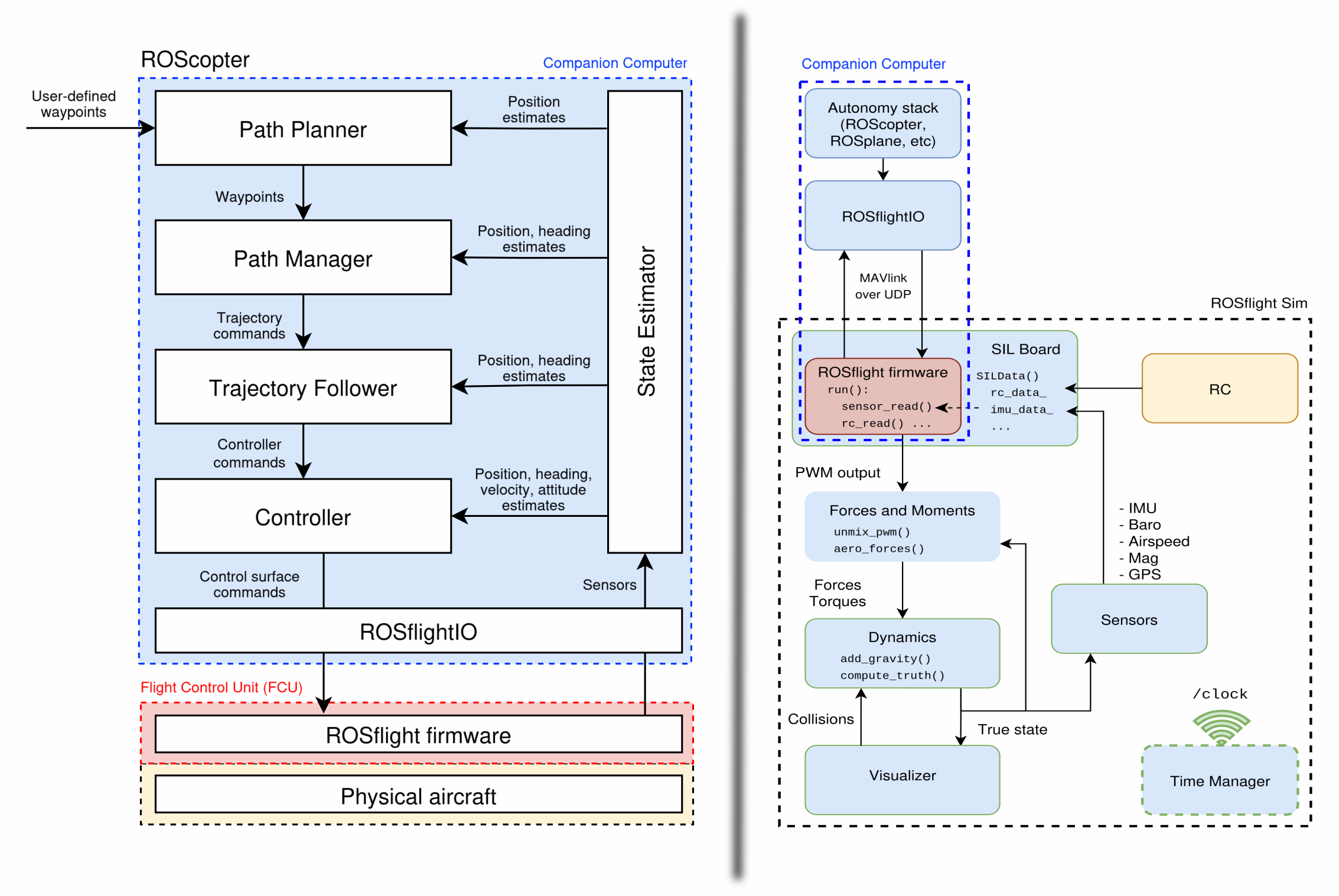}%
    }
    \caption{The ROScopter high-level autonomy stack (left), and software-in-the-loop coordination (right). This figure is adapted from \cite{moore2026roscopter, moore2025rosflight2}.}
    \label{fig:rosflight_architecture}
\end{figure}

The ROSflight autopilot uses the common paradigm of splitting computation between a FCU and a separate onboard computer known as the companion computer. The FCU handles real-time tasks such as low-level sensor aggregation and actuator commands, while the companion computer performs high-level control and planning. Computation on the companion computer is performed within ROS~2. ROS~2 is a middleware that organizes software into nodes, which communicate through topics, services, actions, and parameters \cite{macenski2022ros2}. ROS~2 also helps facilitate a MAVLink connection with the FCU. The ROSflight autopilot can also connect to an optional ground station computer, allowing operators to specify high-level goals for the companion computer, manage system parameters, and monitor downlinked telemetry in real time.

 Within the ROSflight 2.0 architecture on the companion computer, the \texttt{rosflight\_io} node translates ROS~2 topics and service calls to and from MAVLink messages exchanged with the FCU over a serial connection \cite{moore2025rosflight2}.
In ROSflight 2.0, the companion-computer flight stack is separated into distinct ROS~2 nodes for estimation, control, navigation, and application logic. For multirotors, these nodes form the ROScopter autonomy stack shown in Fig.~\ref{fig:rosflight_architecture}.

Below this at the system level, the FCU interfaces directly with sensors, actuators, and the safety pilot's radio-control (RC) receiver. It also manages firmware state and optionally supports the fast estimation and control loops needed for manual or assisted flight.

\section{Rust and Compile-Time Safety}

Program safety is a multifaceted issue; to frame our discussion on safety and comparisons with firmwares developed in unsafe programming languages, we provide an introduction to two critical kinds of program safety that heavily influenced the design of Veloxity: memory safety and type safety.

Memory safety is the proper control of memory in computer programs to avoid improper or unexpected access and release \cite{brinker2025memorySafety}. Traditional systems programming languages (such as C and C++) provide unrestricted access to system resources and rely on programmer discipline to protect against memory-unsafe behavior. While C++ has notably added memory safety features such as smart pointers, taking advantage of all C++ has to offer frequently requires users to integrate their code with unsafe community code \cite{DeLozier2024cppCommunitySafety}. The unsafe foundations of C++ also limit the ability of post hoc additions to provide complete safety \cite{acm2022cppSmartPointers,Rebert2024}.

\begin{figure}[H]
    \centering
    \makebox[\linewidth][c]{%
        \includegraphics[width=\linewidth]{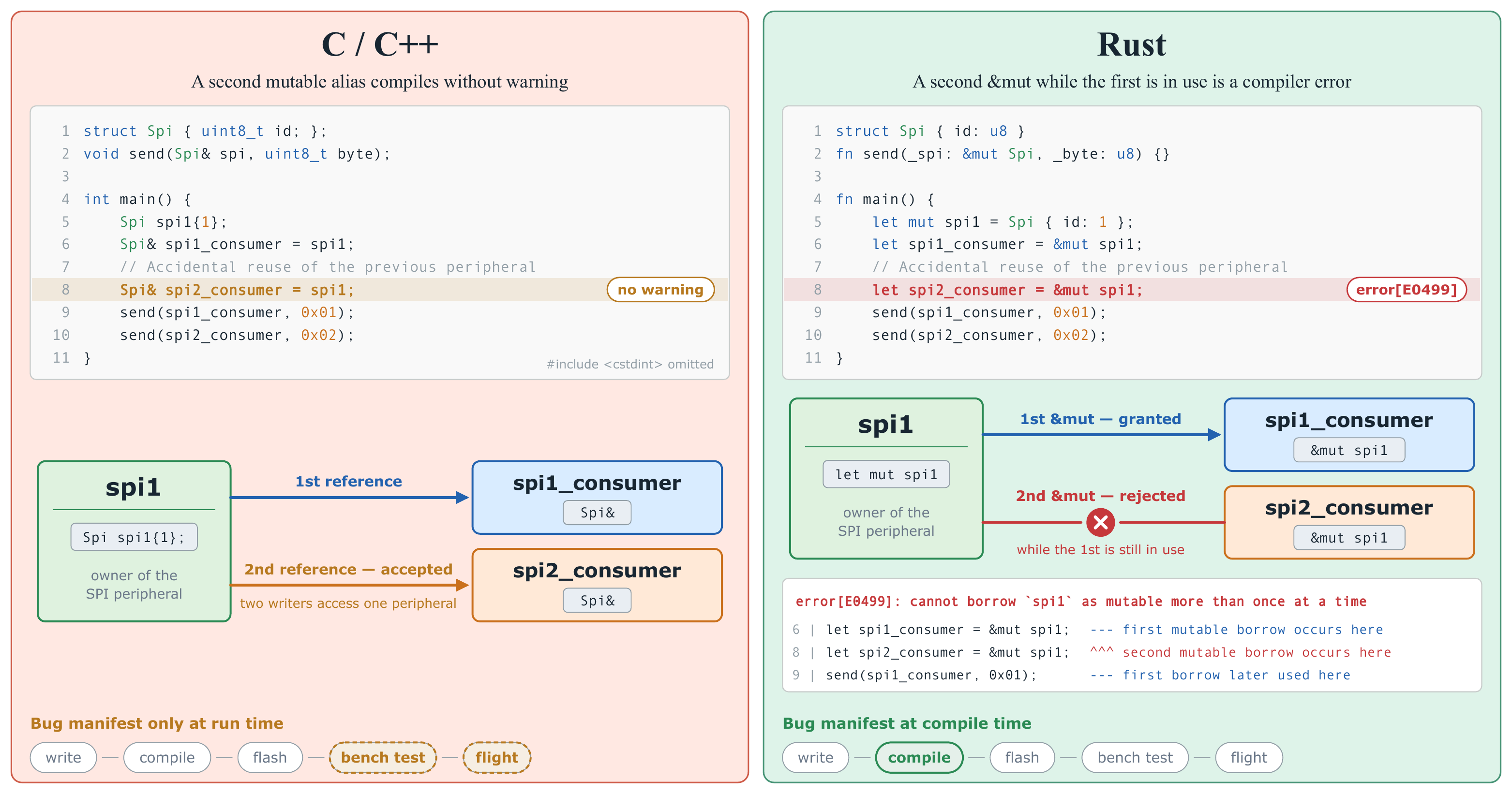}%
    }
    \caption{Rust memory safety, compared to raw pointer access in C and C++.}
    \label{fig:aliasing}
\end{figure}

In Rust's ownership model, memory safety is included by default, and managed through ownership. Ownership can be transferred between variables through move semantics, but unlike in C++, use of a moved-from variable is prohibited until it is reinitialized.
Additionally, beyond the ownership-transfer semantics familiar from C++, Rust's ownership and borrowing system enforces either exclusive mutable access or shared immutable access via ordinary reference to a given value at any point in time \cite{rustbook2026borrowing}.
Figure~\ref{fig:aliasing} outlines the difference between references in C and C++ and in Rust. In this example, a stand-in serial peripheral interface (SPI) is accidentally mutably borrowed twice with overlapping usage. Rust detects the potential for conflicting access and reports an error at compile time. C and C++ allow multiple pointers or references to modify the same object, potentially leading to corrupted state (or data races in unsynchronized concurrency cases). Many such errors in C and C++ would only be observable intermittently at runtime.

We acknowledge approaches to memory safety relying on garbage collection have been successfully implemented with languages such as Python, Java, and Go. While garbage collection can avoid some of the pitfalls incurred from manually managed memory, it comes with a significant performance penalty, especially for memory-constrained systems \cite{Cai2022}. Conversely, Rust makes an excellent choice for memory-safe embedded systems programming due to its low-cost safety guarantees enforced at compile time. To enforce memory safety with minimal runtime overhead, the Rust compiler uses its memory ownership model to perform checks on the validity of memory in advance, eliminating the need for garbage collection.

The Veloxity architecture is also informed by Rust's type safety enforcement. Here, type safety is defined as the guarantee that a program will never misinterpret a value as being of a type for which its behavior is not well defined \cite{defineTypeSafety,weiss2021oxide}. Many systems-level languages (particularly C and C++) allow for flexibility in type usage and interpretation. This flexibility can lead to unexpected behavior, such as unexpected castings. Rust prevents these errors by rigorously checking user-defined types at compile time, requiring that all casts be explicit, and providing types with separate null representations that must be explicitly handled.

We note that in some cases low-level hardware access requires code that cannot be evaluated for safety by the compiler. In cases such as these Rust requires that code be wrapped in \texttt{unsafe} blocks so that it can be carefully audited. Veloxity confines unsafe code to device drivers, as well as a limited selection of supporting structures, such as the simulation's sensor foreign function interface (FFI). The entire control stack maintains Rust's safety guarantees.

\section{Veloxity Architecture}

\begin{figure}[H]
    \centering
    \makebox[\linewidth][c]{%
        \includegraphics[width=\linewidth]{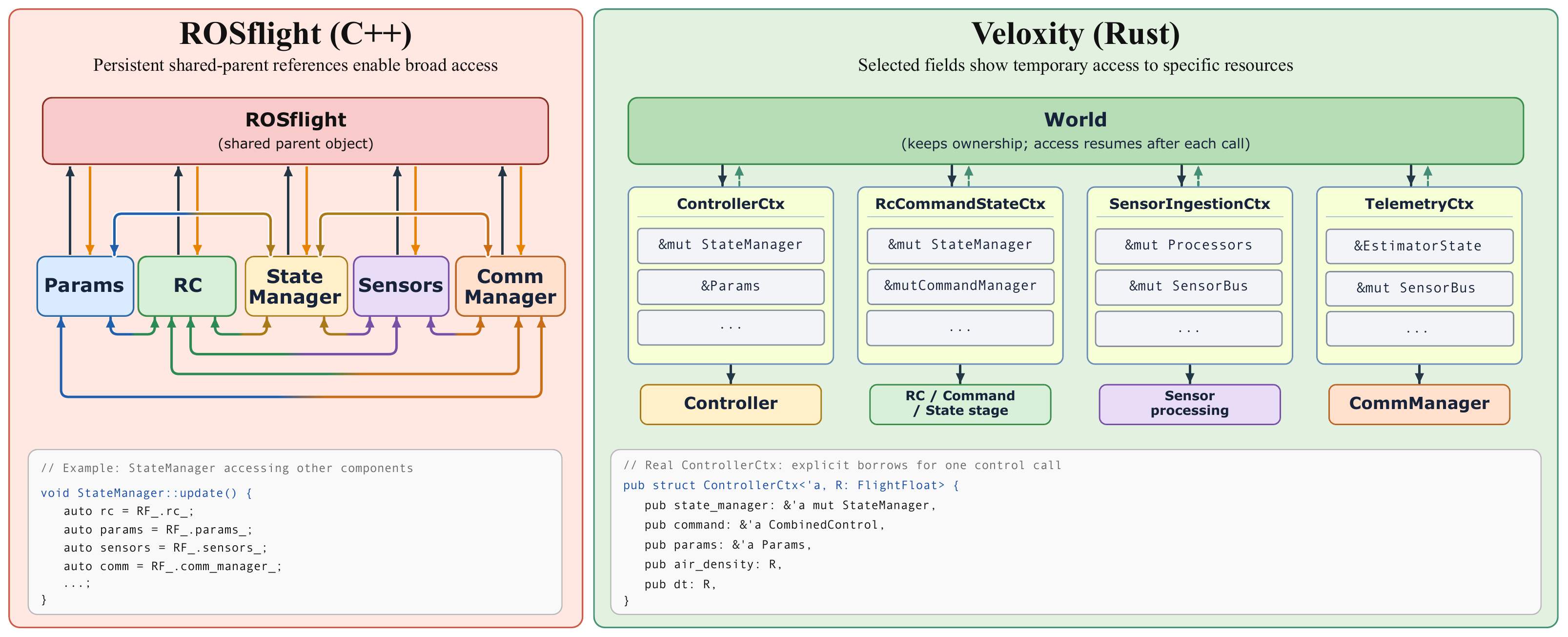}%
    }
    \caption{Shared access in ROSflight versus context-based access in Veloxity.}
    \label{fig:context_access}
\end{figure}

While Veloxity retains the same general modular structure of the ROSflight C++ firmware, the implementation under the hood has been revised to leverage the memory and type safety features of Rust. 
Unlike the ROSflight firmware counterpart, Veloxity's architecture revolves around a strongly typed structure, \texttt{World}, that owns the long-lived resources shared across the firmware. These resources include (but are not limited to) sensor processors, the state estimator, the flight controller, communication systems, system configuration, sensor measurement storage, and the firmware state machine. Rather than allowing individual subsystems to retain unrestricted access to these resources, the \texttt{World} governs when and where resources are temporarily borrowed for specific operations. Through this controlled access, safety guarantees for memory can be enforced by Rust's type system and ownership model at compile time rather than relying on programmer convention alone. Here we highlight three concrete examples of improvements over the ROSflight firmware counterpart derived from the aforementioned architecture:

First, Veloxity implements shared state through contexts. Contexts are defined in this paper as narrowly scoped groups of references that compile together Veloxity's flight-system resources and temporarily expose them for use. Individual processing stages within the core operate through short-lived context structures containing references with move semantics. During the lifetime of a context, only portions of \texttt{World} that need to be exposed for a given operation are accessible, leaving all other parts of \texttt{World} inaccessible. Whereas the ROSflight firmware implementation exposes its major subsystems through a shared parent structure, Rust ownership and borrowing can enforce exclusive mutable access to any given component across the entire flight stack at compile time without sacrificing the architectural reachability of any component from any other. Figure~\ref{fig:context_access} contrasts the two approaches.

\begin{figure}[H]
    \centering
    \makebox[\linewidth][c]{%
        \includegraphics[width=\linewidth]{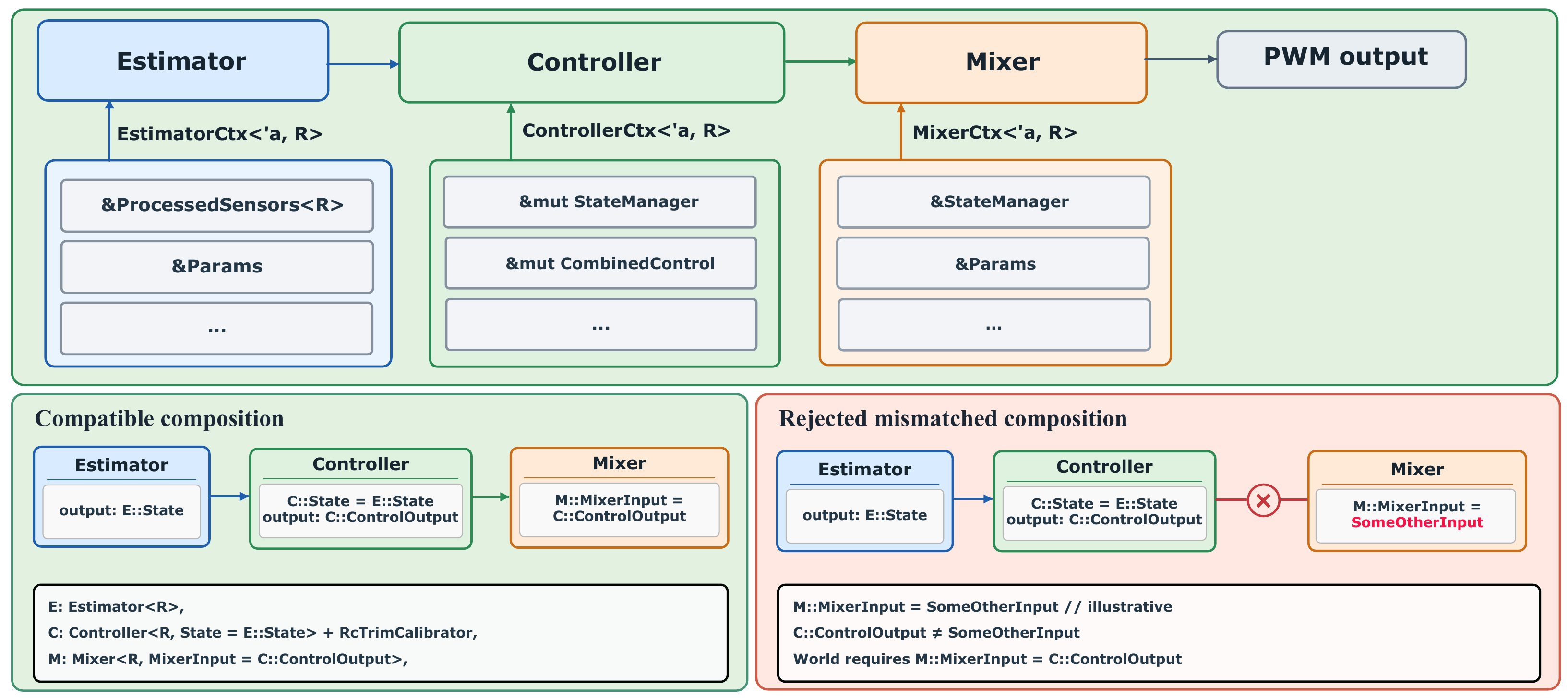}%
    }
    \caption{Compile-time compatibility enforced through associated types in the control pipeline.}
    \label{fig:trait_connectivity}
\end{figure}

Second, Veloxity defines rigorous, statically checked interfaces at subsystem boundaries, using Rust's trait system to encode compatibility directly into the type system. Traits and associated types can require, for example, that a controller interface accept exactly the state type produced by an estimator interface and that a mixer interface accept exactly the control output produced by a controller interface (Fig.~\ref{fig:trait_connectivity}). This means an incompatible composition cannot be instantiated by design, but multiple tracks of estimator, controller, and mixer combinations within their own agreed types are possible. Safer data interfaces even extend into hardware drivers, bringing checked abstractions into low-level buffer access. In Veloxity, these constraints compose naturally through trait bounds, allowing the compiler to verify entire chains of subsystem compatibility rather than relying on repeated casts, assertions, or programmer-maintained assumptions at individual boundaries. We note modern C++ can achieve a similar compatibility through the use of templates, concepts, wrapper types, and assertions, but that doing so requires higher maintenance by the programmer, more complex syntax, and predefining assumptions that must be checked at boundaries manually.

\begin{figure}[H]
    \centering
    \makebox[\linewidth][c]{%
        \includegraphics[width=\linewidth]{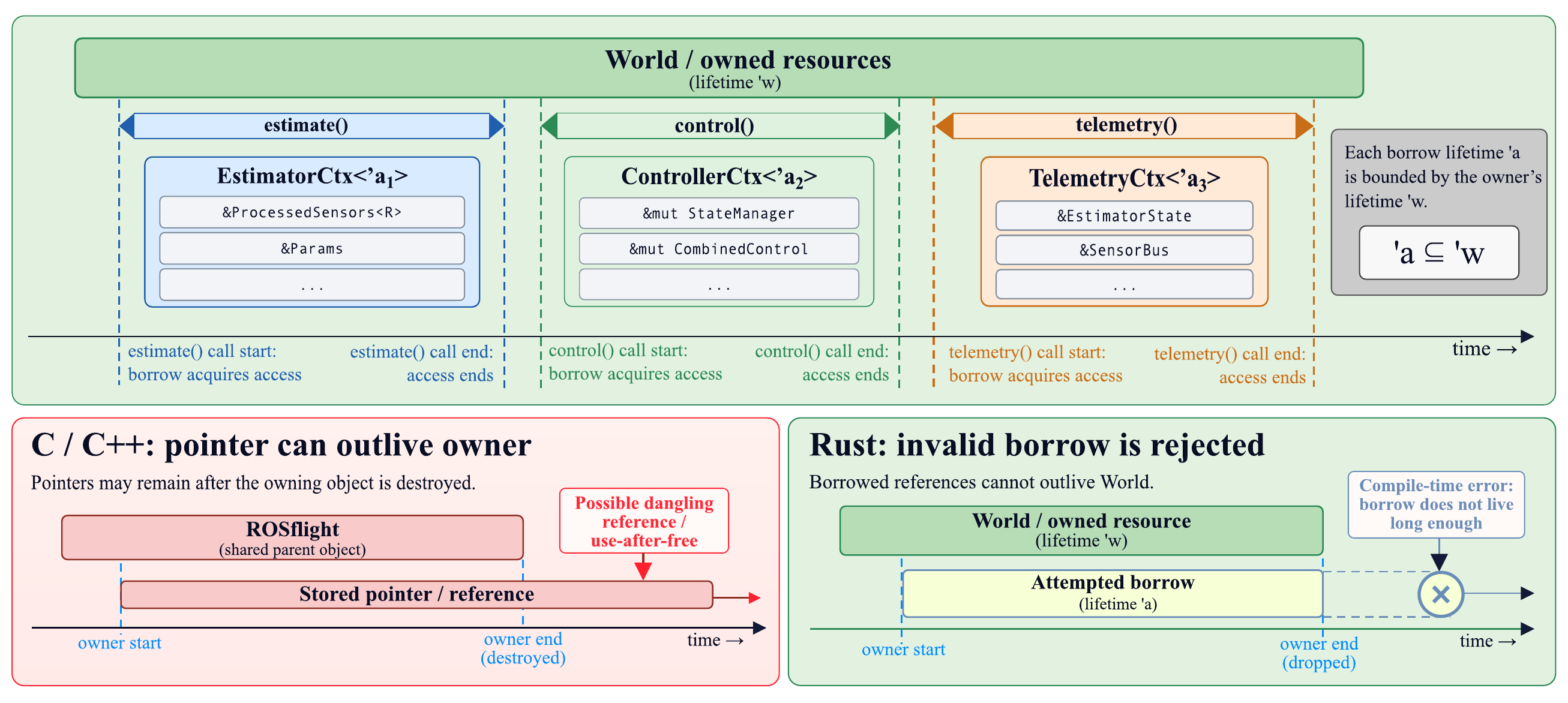}%
    }
    \caption{Diagram depicting how Rust's lifetime system prevents dangling references.}
    \label{fig:lifetimes}
\end{figure}

Third, Rust's lifetime system extends the aforementioned compile-time guarantees dealing with exclusive memory access and memory validity to the references themselves. References passed between Veloxity subsystems carry compile-time lifetime relationships that prevent them from outliving the resources they reference (Fig.~\ref{fig:lifetimes}), eliminating classes of dangling-reference and use-after-free errors without requiring runtime checks. Modern C++ can strongly constrain interfaces using references, smart pointers, spans, concepts, and static analysis, but its type system generally does not encode or prove the lifetime relationship between a borrowed reference and the object from which it originated. Consequently, even a correctly typed C++ interface may still admit a reference whose underlying storage has expired, whereas the equivalent safe Rust program is rejected during compilation. This allows Veloxity to express not only what data may cross an interface and who may mutate it, but also how long that access is valid. 

The distinction is even more pronounced in C, which provides neither language-level ownership abstractions nor mechanisms for enforcing reference lifetimes. Memory validity in C must be managed through programmer discipline and external analysis, leaving dangling pointers, use-after-free errors, and conflicting memory access largely unchecked. With so many quick paths to unsafe code in embedded environments where performance cannot afford to fail, being able to enforce safety by default makes Rust a natural choice.

\subsection{Flight-Stack Core}

Veloxity further distinguishes itself from the ROSflight firmware implementation through the organization of communication and data exchange within the flight stack. ROSflight already uses strongly typed structures, references, and function arguments to pass information between subsystems. However, several operations are handled directly through callbacks that invoke other subsystem methods. For example, parameter-setting callbacks in the ROSflight communication manager directly access and modify the parameter subsystem. This pattern can conflict with Rust's borrowing rules when callbacks require mutable access to subsystem state that is already borrowed by the calling function, preventing a direct translation of the implementation into safe Rust.

Adapting to restrictions such as this, Veloxity separates the reception of certain messages from their processing through statically typed event queues. Incoming communication messages are decoded into protocol-independent structures, from which events such as \texttt{ParamSetRequested} and \texttt{OffboardControlRequested} are generated. These events are stored by value and subsequently consumed by the appropriate processing stages, which may generate additional events or communication responses. This approach makes the flow of requests and responses explicit while allowing processing stages to operate independently of the communication protocol that originated them. Time-critical operations, including estimation, control, and actuator mixing, continue to use direct, typed function calls rather than passing every intermediate result through an event queue.

Veloxity implements these event queues using fixed-capacity buffers provided by the \texttt{heapless} crate (a Rust library package) \cite{aparicio2026heapless}. Each queue is parameterized by the type of message it accepts and its maximum capacity, preserving message types across processing boundaries while bounding memory consumption without dynamic allocation. Rust's type system enforces compatible queue operations, while ownership and borrowing rules constrain how producers and consumers access the underlying storage. These guarantees do not, however, eliminate the need for runtime validation or ensure that every message is delivered: a full queue must still be handled, and Veloxity may log and discard events when capacity is exhausted. Modern C++ can also implement typed, statically allocated queues, making this primarily an architectural distinction rather than a capability exclusive to Rust. Veloxity's improvement lies in combining bounded event-based communication with compiler-enforced access constraints and explicit separation of processing responsibilities. This separation also allows noncritical service operations to be deferred without interrupting higher-priority flight-control processing, as discussed in the following section.

\subsection{Core Scheduling}

\begin{figure}[H]
    \centering
    \makebox[\linewidth][c]{%
        \includegraphics[width=\linewidth]{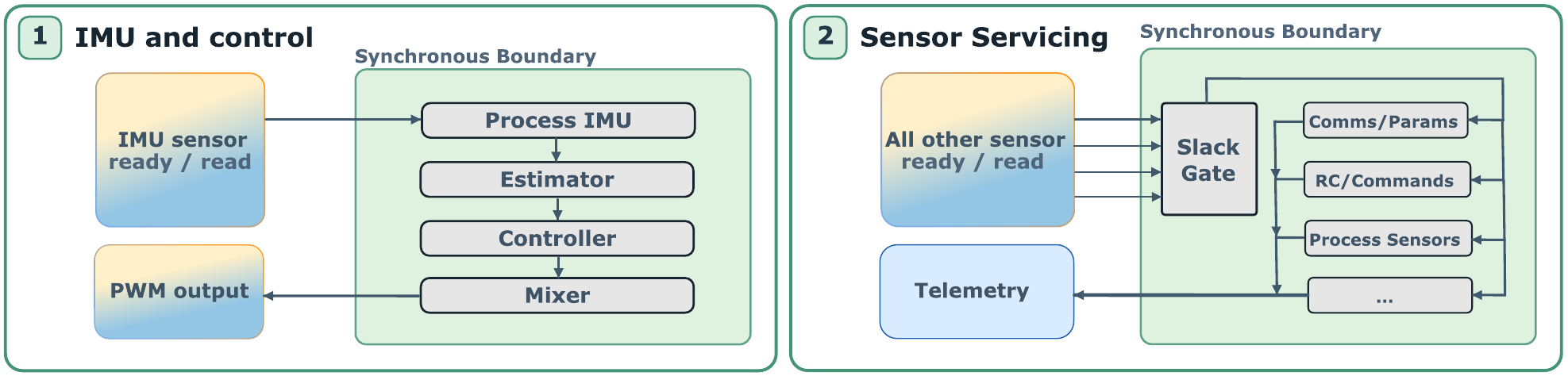}%
    }
    \caption{Veloxity core scheduling. Timing-critical control events are given priority over general service calls.}
    \label{fig:core_scheduling}
\end{figure}

Veloxity has been designed to prioritize estimation and control, and succeeds at minimizing jitter and nondeterministic delays.
To ensure that flight-critical computations are always prioritized, the Veloxity main loop selects between three active paths in descending priority: core sensor processing, flight-control updates, and service processing (Fig.~\ref{fig:core_scheduling}).
Each path through the main loop first borrows components from the \texttt{World} structure as needed (ensuring safe, non-duplicated memory access), performs computation, and returns those components back to \texttt{World}.
To accommodate bursts in processing demand (e.g. multiple sensors coming due at once) without dynamically allocating memory and without violating safe Rust rules, event queues built with the \texttt{heapless} crate retain service requests, allowing the prioritized loop to defer less critical work or processing until higher priority tasks have been allowed to complete.

At the lowest level, the interrupt-based sensor acquisition processing mirrors the ROSflight firmware in functionality.
Veloxity uses the Embassy framework \cite{embassy2026} to run asynchronous peripheral tasks on interrupt executors, with IMU acquisition assigned a higher priority than communication and lower-rate sensing.

\subsection{Integration with ROS~2 Simulation Stack}

ROS~2 has become a common language around which the robotics community has developed a vast library of out-of-the-box tooling. This tooling supports rapid development, and assists reproducibility in the form of package management and distribution. ROSflight was originally developed on the backbone of ROS~2 so as to share these advantages. Integrating Veloxity with ROSflight, we bridge the gap between a memory-safe backbone and the extensive high-level capabilities offered by the ROS~2 community.

\begin{figure}[H]
    \centering
    \makebox[\linewidth][c]{%
        \includegraphics[width=\linewidth]{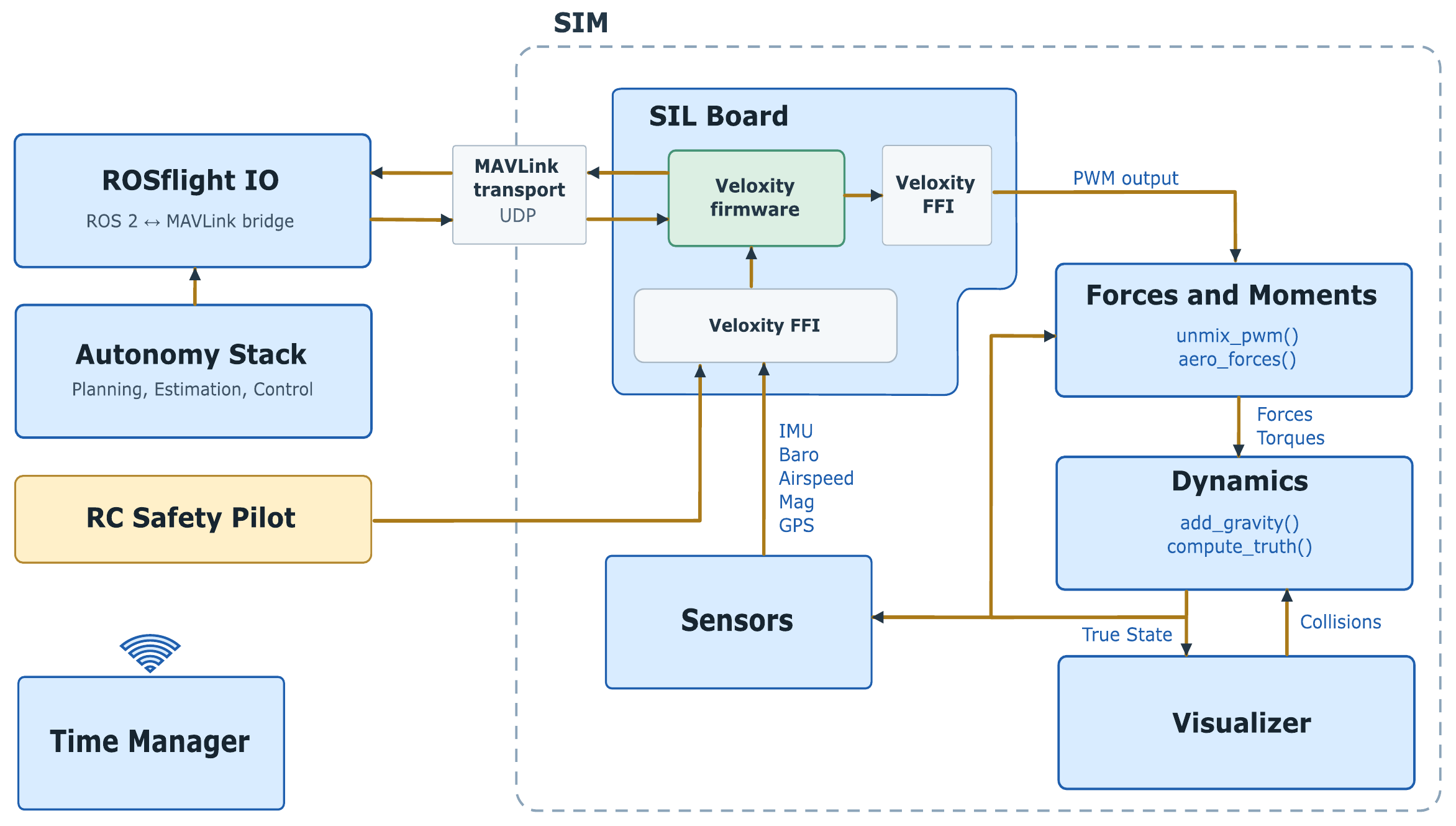}%
    }
    \caption{Software-in-the-loop architecture with Veloxity.}
    \label{fig:sim_architecture}
\end{figure}

Our complete integration requires preserving two distinct connections: the link used by the autonomy stack to communicate with the firmware, and the implementation of the board interface through which the firmware receives measurements and produces actuator commands for software-in-the-loop simulation \cite{moore2025rosflight2}.
The first carries MAVLink messages through \texttt{rosflight\_io}, using serial communication when connected to hardware, and User Datagram Protocol (UDP) when configured for SIL simulation. Nothing is changed for this connection when swapping between the ROSflight firmware and Veloxity.
The second connection is the board's physical sensor interface: only in SIL does it require an extra interface at the boundary. At said boundary, a C++ ROS~2 adapter (labeled Veloxity FFI in Fig.~\ref{fig:sim_architecture}) receives the simulator's published sensor data and passes it to a Rust implementation of the board interface.

The adapter subscribes to the existing sensor topics and converts their messages into the measurement types expected by Veloxity's board interface, leaving sensor generation in the ROSflight simulator.
The flight-control core therefore receives simulated measurements through the same interface used for hardware measurements, without depending on ROS~2 message types.
Actuator commands return through the adapter to the simulator's existing pulse-width-modulation (PWM) topic, closing the simulated control loop without routing raw sensor input through \texttt{rosflight\_io}.
ROSflight continues to publish sensor and RC data, invoke the \texttt{sil\_board/run} service, consume actuator commands from \texttt{sim/pwm\_output} node, and communicate through the existing \texttt{rosflight\_io} node as is standard for the ROSflight architecture.

To connect the C++ adapter to Rust, the Veloxity simulation runtime is compiled as a Rust static library and exposed through a narrow C application binary interface.
A C++ ROS~2 adapter manages a firmware handle and transfers data by copying a fixed-layout sensor record into Rust and a fixed-size array of actuator outputs back into C++.
Matching C declarations and Rust \texttt{repr(C)} structures establish a compatible layout, while \texttt{extern "C"} functions provide the calling convention across the language boundary.
The firmware itself remains Rust-owned and executes on a separate worker thread. The C++ adapter holds a handle rather than accessing its internal structures and releases it through a Rust destruction function on ROS~2 node shutdown.
This design confines raw-pointer handling to a small interface, but does not extend Rust's compile-time checks into C++: compatible data layouts, valid pointers and correct lifetime handling remain obligations of the integration.

Within the worker thread running Veloxity, we have configured the firmware to run continuously under the core scheduler rather than executing one control step per ROS~2 service call. This mirrors the unsynchronized reality experienced in hardware.
Preserving both connections allows ROSflight firmware and Veloxity to be compared using the same simulator and autonomy interfaces, with \texttt{rosflight\_io} left unmodified.

\section{Experiments and Comparison with ROSflight}

In this section, we present experimental comparisons between Veloxity and the ROSflight firmware written in C++.
We evaluate four criteria: deployment/resource footprint, timing, algorithmic alignment with the ROSflight firmware (compared through simulations performed under equivalent noise profiles), and hardware-demonstrated results.
Under the evaluated configurations, Veloxity performs comparably in both simulation and outdoor flight, showcasing successful operation on the STM32H7 flight controller and real-time integration with the ROScopter autonomy stack.

ROSflight 2.0 provides first-class support for the STM32H7 microcontroller family, which is integrated into the commercially available Pixracer Pro FCU. Veloxity offers the same first-class support and out-of-the-box functionality. As such, the STM32H7 platform will be the basis of comparison for baseline timing tests, and binary size between ROSflight firmware and Veloxity.

\subsection{STM32H7 Deployment and Resource Footprint}

Embedded flight controllers operate under strict memory constraints, making firmware resource requirements an important consideration when evaluating alternative implementations. To qualify as a practical alternative to existing C and C++ firmware, the benefits of Rust's compile-time memory-safety guarantees must be achieved without introducing prohibitive resource overhead. We therefore compare the release-build flash footprints of Veloxity and the original ROSflight firmware on the STM32H743II microcontroller used by the Pixracer Pro.

\begin{table}[hbt!]
\caption{\label{tab:firmware_size} Release-build resource footprint on the Pixracer Pro's STM32H743II. All sizes are in bytes.}
\centering
\small
\setlength{\tabcolsep}{4pt}
\begin{adjustbox}{max width=\textwidth}
\begin{tabular}{@{}lcc@{}}
\toprule
Measurement & ROSflight & Veloxity \\
\midrule
Raw flash image & 415,640 & \textbf{412,352} \\
Read-only code/data & \textbf{271,336} & 396,672 \\
\bottomrule
\end{tabular}
\end{adjustbox}
\end{table}

Veloxity's raw flash image is 3,288 bytes (0.79\%) smaller than the ROSflight firmware baseline (Table~\ref{tab:firmware_size}). This demonstrates Veloxity can be deployed on the same hardware without increasing overall flash requirements. Though the two implementations occupy nearly the same amount of flash, their distributions of executable code, read-only data, and initialized writable data differ. Similar differences in the memory-section composition of C and Rust firmware have been documented in prior work, reflecting the effects of compiler behavior and implementation choices \cite{thapa2026embedded}.

\subsection{Companion-Computer Control and Communication Timing}

ROSflight can be configured to close every loop on the companion computer, offloading control responsibilities from the firmware. Commands from the companion computer in this configuration bypass firmware controllers and pass directly to the mixer, while the firmware retains sensor acquisition, actuator mixing, and safety responsibilities. ROSflight 2.0 demonstrates this arrangement at 400~Hz \cite{moore2025rosflight2}, and ROScopter uses it for waypoint following \cite{moore2026roscopter}. Embedded control depends mainly on timely execution within the firmware, whereas companion-based control depends on timely delivery of commands and sensor data between the two computers. We repeat the command-echo test from the ROSflight 2.0 hardware evaluation \cite{moore2025rosflight2} with both firmwares running on the same Pixracer Pro and Raspberry Pi 5 hardware. The companion computer publishes offboard commands at 400~Hz through \texttt{rosflight\_io}, and the firmware echoes each received command back.

Timing starts in the \texttt{rosflight\_io} command callback and ends when the echo is decoded, so each round-trip time includes MAVLink serialization, serial transport, firmware parsing, any scheduling delay in the firmware, and return transmission time. It excludes ROS~2 delivery before the callback and does not establish when the corresponding actuator output is applied. We use the Raspberry Pi 5 and Pixracer Pro USB configuration from the original test (configuration~1), at a configured baud rate of 921,600. Both firmwares use the same command rate, serial and telemetry configuration, and parameter set. For each firmware, commands are published for 180~s after a 5~s warm-up, and the round-trip time of every returned echo is recorded.

\begin{figure}[H]
    \centering
    \makebox[\linewidth][c]{%
        \includegraphics[width=\linewidth]{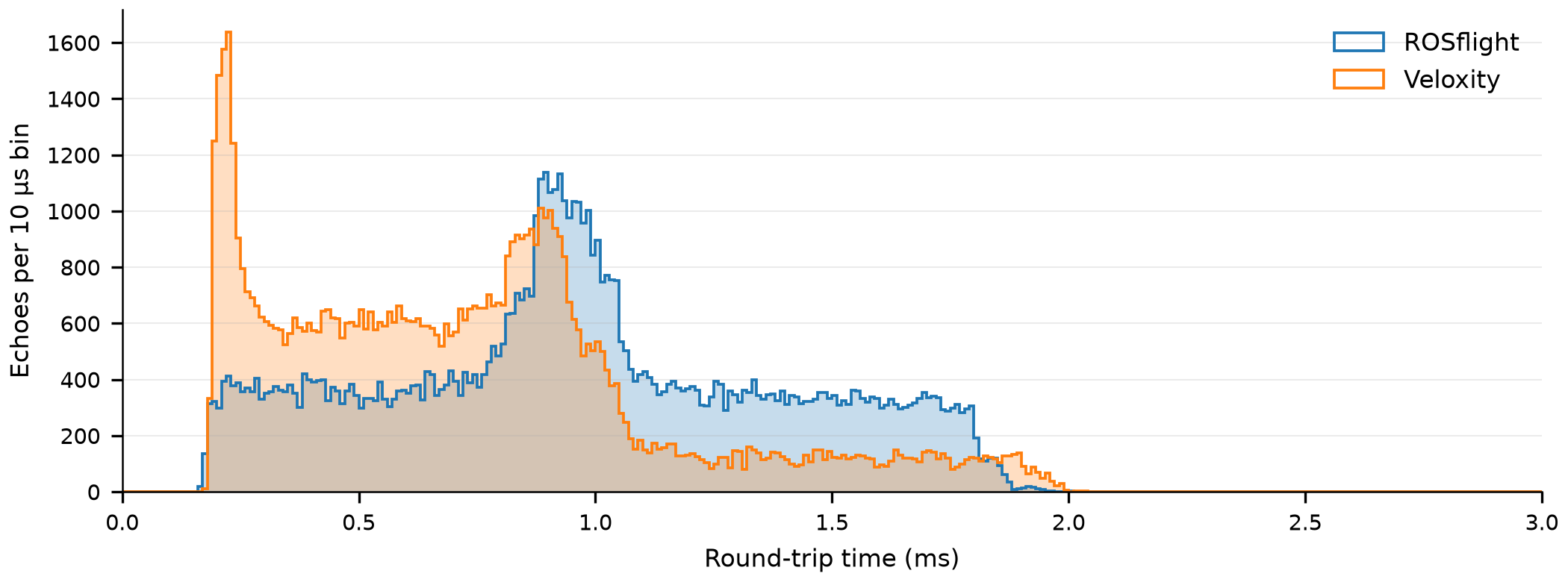}%
    }
    \caption{Round-trip times at or below 3~ms for 400~Hz offboard commands.}
    \label{fig:latency_distribution}
\end{figure}

Figure~\ref{fig:latency_distribution} shows the distribution of round-trip times at or below 3~ms, and Table~\ref{tab:rtt_stats} summarizes the first response to each command. Veloxity lowers the mean round-trip time from 0.974~ms to 0.738~ms (approximately 24\%), and also lowers the maximum and 95th percentiles. Minimum and 99th percentile delays remain slightly higher. 

Table~\ref{tab:response_completeness} reports response completeness. Of 71,999 commands published to each firmware, Veloxity echoed every command exactly once. The ROSflight firmware failed to echo five commands, and returned 417 duplicate echoes (which all arrived after 7.5~ms). Those duplicates account for the ROSflight firmware's large delays reaching up to 88.059~ms.

\begin{table}[hbt!]
\caption{\label{tab:rtt_stats} Round-trip times for first responses to 400~Hz offboard commands, in milliseconds. Duplicate echoes are excluded.}
\centering
\begin{adjustbox}{max width=\textwidth}
\begin{tabular}{lrrrrr}
\toprule
Firmware & Mean & Min & Max & 95th pct. & 99th pct. \\
\midrule
ROSflight & 0.974 & \textbf{0.162} & 5.733 & 1.713 & \textbf{1.814} \\
Veloxity & \textbf{0.738} & 0.176 & \textbf{3.415} & \textbf{1.646} & 1.887 \\
\bottomrule
\end{tabular}
\end{adjustbox}
\end{table}

\begin{table}[hbt!]
\caption{\label{tab:response_completeness} Response completeness for the 180~s, 400~Hz command echo test. A missing response is a published command that was never echoed. A duplicate echo is any echo received beyond the first return for the same command.}
\centering
\begin{adjustbox}{max width=\textwidth}
\begin{tabular}{lrr}
\toprule
Measurement & ROSflight & Veloxity \\
\midrule
Published commands & 71,999 & 71,999 \\
Missing responses & 5 & \textbf{0} \\
Duplicate echoes & 417 & \textbf{0} \\
Echoes above 80~ms & 38 & \textbf{0} \\
Maximum delay including duplicates (ms) & 88.059 & \textbf{3.415} \\
\bottomrule
\end{tabular}
\end{adjustbox}
\end{table}

\subsection{Software-in-the-Loop Behavioral Comparison}

Veloxity seeks to change the firmware implementation without changing the mathematical behavior the rest of the ROSflight system expects from the firmware block. We examine this at two levels: numerical agreement of the controller and mixer outputs in isolation, and closed-loop behavior in SIL simulation when each firmware operates through the same ROScopter autonomy stack.

For the numerical comparison, a host-only harness runs the unmodified controller and quad-X mixer of each firmware, without ROS~2 or vehicle dynamics. It applies the same input to both the ROSflight firmware and Veloxity. Table~\ref{tab:numerical_parity} reports results for Veloxity's single-precision \texttt{f32} build, matching the ROSflight firmware single-precision floats. In the \emph{independent} case, each of 10,000 independent firmware instances starts from freshly initialized controller and mixer state. The cases span attitude, rate, throttle, and pass-through commands, angle and rate modes, saturation, armed and disarmed states, and boundary time steps. In the \emph{stateful} case, samples are formed from 100 traces each with 100 consecutive 400~Hz simulated updates, complete with proportional-integral-derivative (PID) control, with state carried from one update to the next across each trace. Together, the independent test checks single updates, while the stateful test checks that differences do not accumulate through state feedback. Veloxity can be built in single or double precision (\texttt{f32} or \texttt{f64}). Here we report the \texttt{f32} build. Against the same ROSflight outputs, the \texttt{f64} build's largest differences are also within single-precision rounding error.

\begin{table}[hbt!]
\caption{\label{tab:numerical_parity} Maximum absolute difference between the controller and mixer outputs of Veloxity's single-precision build and the ROSflight firmware, given identical inputs.}
\centering
\small
\setlength{\tabcolsep}{4pt}
\begin{adjustbox}{max width=\textwidth}
\begin{tabular}{@{}lccc@{}}
\toprule
Comparison & \shortstack{Cases or\\samples} & \shortstack{Controller\\maximum\\difference} & \shortstack{Mixer\\maximum\\difference} \\
\midrule
Independent inputs & 10,000 & $5.970\times10^{-8}$ & $8.941\times10^{-7}$ \\
Stateful samples   & 10,000 & $1.494\times10^{-8}$ & $2.087\times10^{-7}$ \\
\bottomrule
\end{tabular}
\end{adjustbox}
\end{table}

To further test the firmware in a controlled simulation environment, we perform a set of full-stack SIL runs with matched sensor-noise sequences.
The selected mission follows three north-east-down (NED) waypoints, $(0,0,-5)$, $(-20,0,-8)$, and $(-20,20,-5)$, combining horizontal motion with altitude changes.
For each of four runs, the ROSflight firmware and Veloxity share a noise innovation vector (indexed by sensor stream and sample number) to allow firmware differences to be studied without changing the prescribed noise realization. In testing both firmwares, a dedicated feedforward node and a thrust-to-throttle conversion node are added to the default ROScopter graph, and ROSflight's standard quad-x body configuration is chosen.

\begin{table}[hbt!]
\caption{\label{tab:sil_tracking_stats} SIL tracking error of the ROSflight firmware and Veloxity over four matched-noise seed pairs. RMSE is computed between the simulator's true position and the time-parameterized trajectory command over each full run. The minimum equivalence margin is the smallest symmetric margin for which paired two one-sided tests (TOST) establishes equivalence at the 5\% level.}
\centering
\small
\setlength{\tabcolsep}{4pt}
\begin{adjustbox}{max width=\textwidth}
\begin{tabular}{@{}lccccccc@{}}
\toprule
Axis & \shortstack{ROSflight mean\\RMSE (m)} & \shortstack{Veloxity mean\\RMSE (m)} & \shortstack{Mean paired\\difference (cm)} & \shortstack{Paired 90\%\\CI (cm)} & \shortstack{Paired\\$t$-test $p$} & \shortstack{Welch's\\$t$-test $p$} & \shortstack{Minimum\\equivalence\\margin (cm)} \\
\midrule
Horizontal & 1.383 & 1.387 & $+0.44$ & $[-0.38,\ +1.27]$ & 0.294 & 0.992 & 1.27 \\
Vertical   & 1.405 & 1.405 & $+0.00$ & $[-1.22,\ +1.22]$ & 0.998 & 1.000 & 1.22 \\
\bottomrule
\end{tabular}
\end{adjustbox}
\end{table}

\begin{figure}[H]
    \centering
    \makebox[\linewidth][c]{%
        \includegraphics[width=\linewidth]{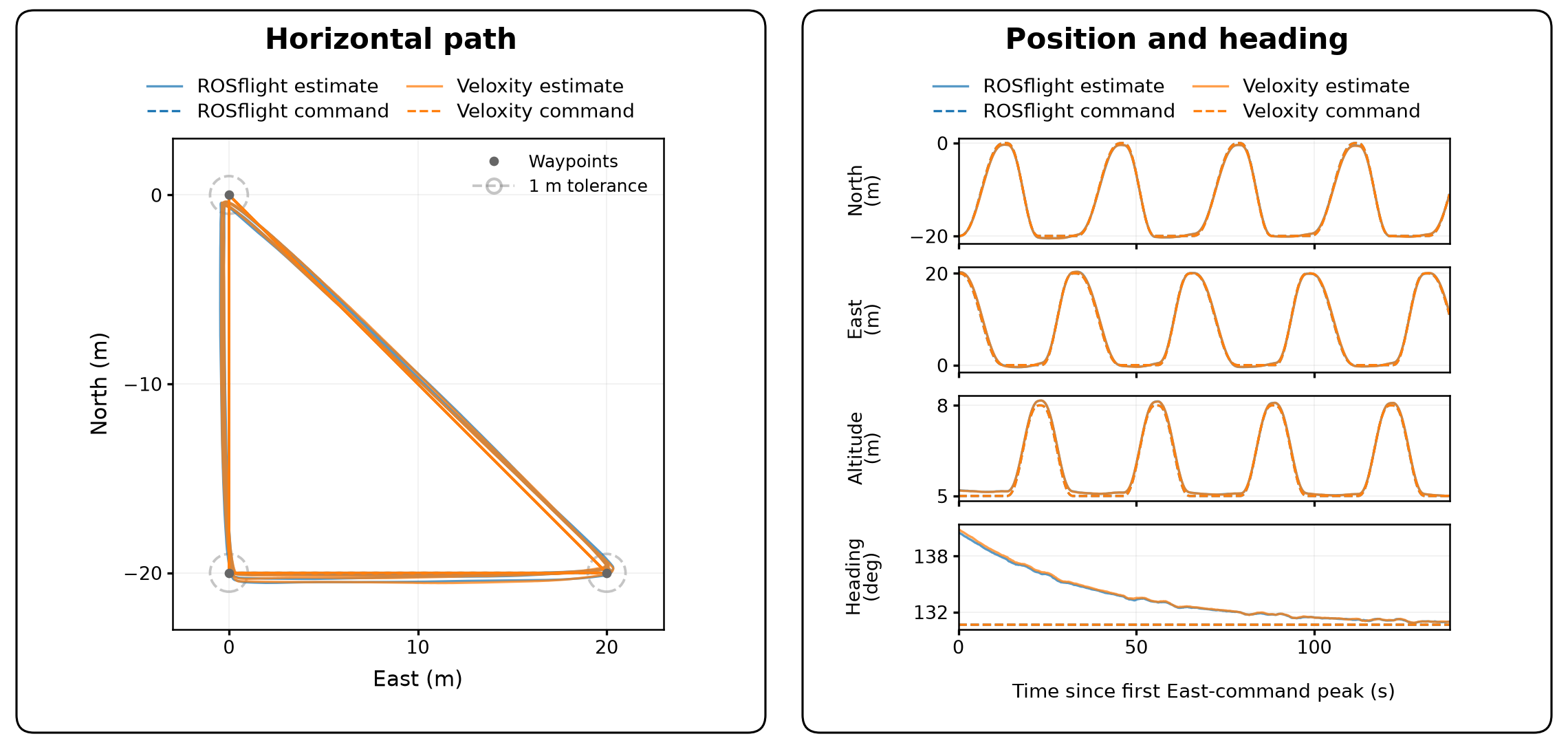}%
    }
    \caption{Software-in-the-loop comparison of ROSflight firmware and
      Veloxity over a 137.98~s interval.}
    \label{fig:sil_comparison}
\end{figure}

Figure~\ref{fig:sil_comparison} shows one of the four matched-noise seed pairs, and Table~\ref{tab:sil_tracking_stats} summarizes the tracking root-mean-square error (RMSE) across all four pairs. Neither a paired $t$-tests nor an unpaired Welch's $t$-test detects a statistically significant difference in horizontal or vertical RMSE at the 5\% level. Paired TOST establish equivalence of the mean RMSE for any margin of at least 1.3~cm (rounded). Slight drift in waypoint timing is attributed to thread scheduling in the multithreaded SIL and, potentially, to the FFI layer that passes ROS~2 sensor data to Veloxity. Throughout the tests, the noise is verified to be identical between firmwares across all timesteps.

\subsection{Outdoor Flight Evaluation}

The outdoor flight test exercises Veloxity's embedded flight controller during waypoint following on a quadrotor, alongside two established reference platforms: PX4 and ROSflight C++ firmware. For these references we use the ROSflight and PX4 flight data supplied from the ROScopter study \cite{moore2026roscopter}. That study concluded that ROScopter provides similar basic waypoint-following performance to PX4 in a lean, modular research platform, without PX4's full feature set.

\begin{figure}[H]
    \centering
    \makebox[\linewidth][c]{%
        \includegraphics[width=\linewidth]{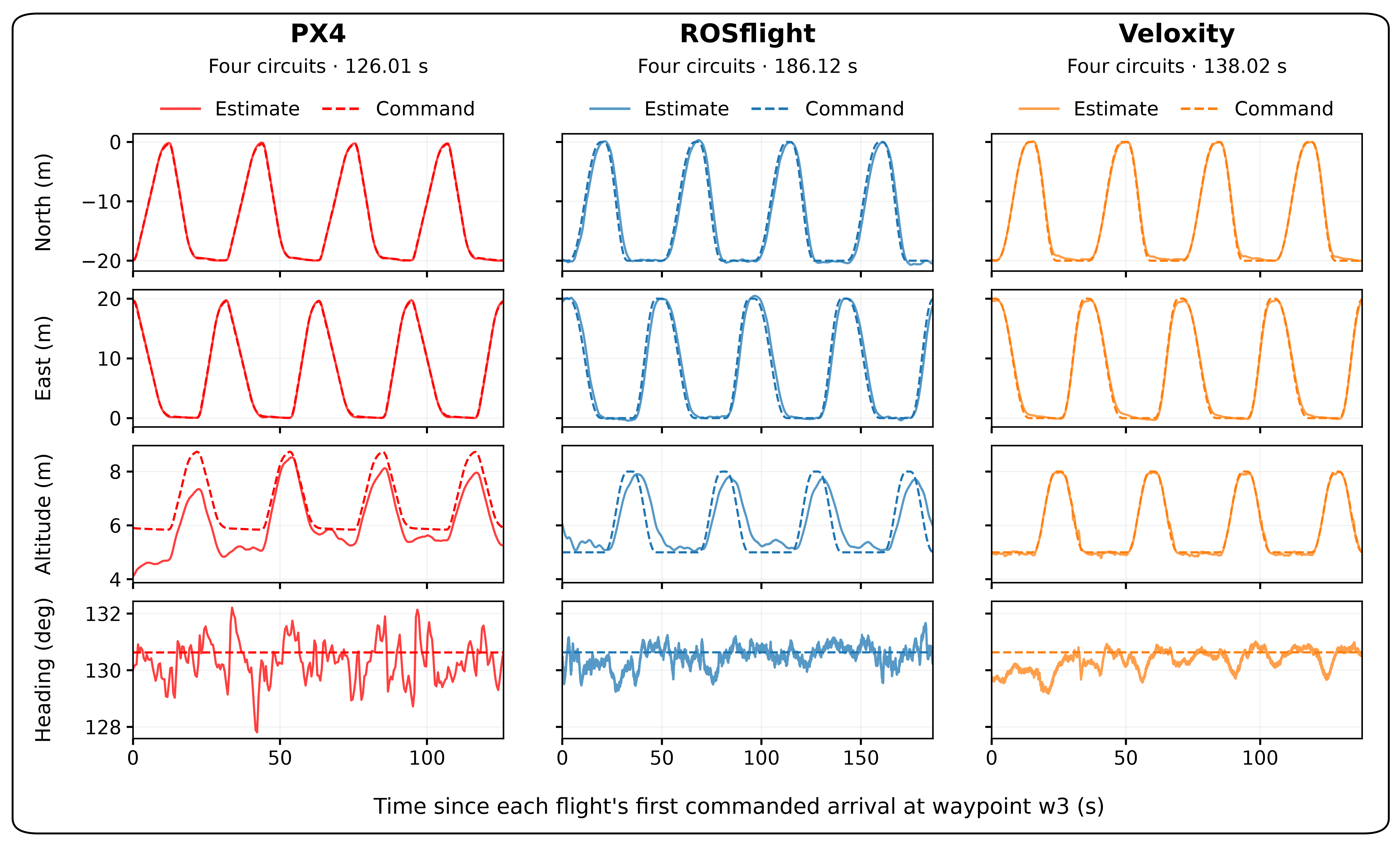}%
    }
    \caption{ Four waypoint circuits from the PX4, ROSflight firmware and Veloxity outdoor flights.}
    \label{fig:outdoor_flights}
\end{figure}

\begin{figure}[H]
    \centering
    \makebox[\linewidth][c]{%
        \includegraphics[width=\linewidth]{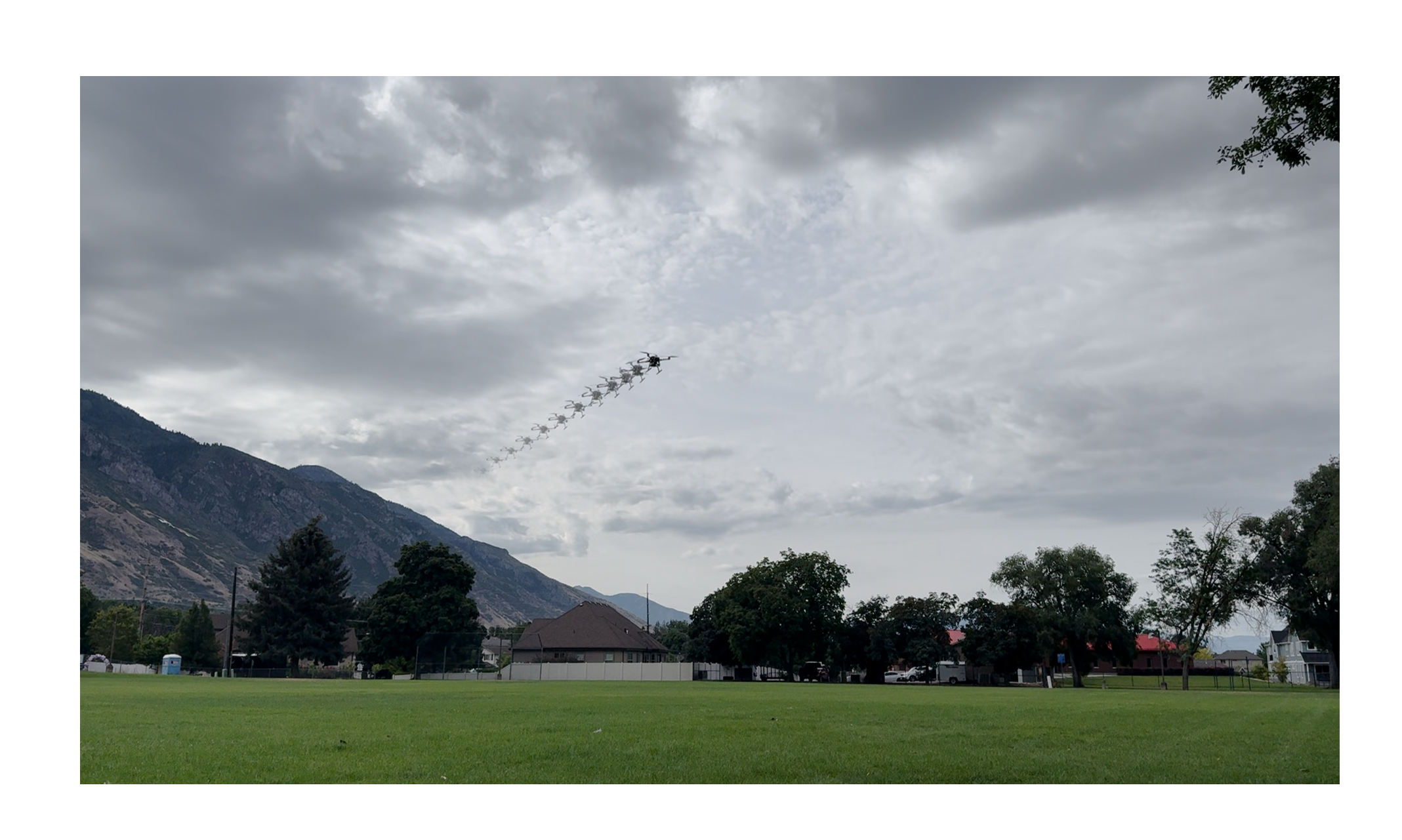}%
    }
    \caption{Holybro X650 quadrotor running Veloxity (running on a Pixracer Pro FCU) and the ROScopter autonomy stack (running on a Raspberry Pi 5).}
    \label{fig:outdoor_vehicle}
\end{figure}

The Veloxity flight follows the same waypoint mission as the published ROSflight firmware flight \cite{moore2026roscopter}: three NED waypoints, $(0, 0, -5)$, $(-20, 0, -8)$, and $(-20, 20, -5)$~m, with a fixed $130^\circ$ heading. We note that for our outdoor test using Veloxity we use the same configuration tested in the SIL simulation section. The published PX4 commanded altitudes run about 0.8~m above those of the ROScopter mission (Fig.~\ref{fig:outdoor_flights}). All three flights use a Pixracer Pro FCU and a Holybro X650 frame (Fig.~\ref{fig:outdoor_vehicle}), and the ROSflight firmware and Veloxity flights share a Raspberry Pi 5 companion computer. We enable Veloxity's full embedded angle-control path. The companion computer closes the position-control loop, while the FCU performs attitude estimation, roll and pitch angle control, yaw-rate control, mixing and actuator output, alongside sensor processing, communication, and safety checks. Veloxity schedules control updates at 400~Hz, a nominal period of 2.5~ms. The flight-controller timestamps on the recorded IMU messages confirm an onboard rate of 399.8~Hz, matching the ROSflight C++ firmware flight's 399.7~Hz. PX4 v1.15 defaults to a 400~Hz inner-loop rate limit. Figure~\ref{fig:outdoor_flights} shows four waypoint circuits from each flight.

Table~\ref{tab:outdoor_tracking} reports tracking error over the four circuits shown in Fig.~\ref{fig:outdoor_flights}. We use the metric of Moore et al.~\cite{moore2026roscopter}: position error is each autopilot's estimated position minus the closest point on the ideal straight-line path between waypoints. We report RMSE in North, East, and Down, with total RMSE as the square root of the sum of their squares. Applied to the study's original windows, our implementation reproduces its published totals of 0.533~m for ROSflight and 0.469~m for PX4. However, the study scored the ROSflight flight over its whole recording and PX4 only after its altitude transient ($t \geq$ 63~s). We instead apply one window to all three flights: from each flight's first commanded arrival at a predefined waypoint, over the next twelve waypoint legs (four complete circuits). This forms the longest span that all three recordings contain from a common starting waypoint. Because this path metric does not depend on timing, a vehicle that follows the path while lagging its time-parameterized command is not penalized, which makes it suitable for flights flown at different speeds.

\begin{table}[hbt!]
\caption{\label{tab:outdoor_tracking} Outdoor waypoint-tracking error over the four circuits shown in Fig.~\ref{fig:outdoor_flights}. Error is the distance from each autopilot's estimated position to the nearest point on the straight-line path between the published mission waypoints \cite{moore2026roscopter}. Total RMSE is the root-sum-square of the three axis RMSEs.}
\centering
\small
\setlength{\tabcolsep}{4pt}
\begin{adjustbox}{max width=\textwidth}
\begin{tabular}{@{}lcccc@{}}
\toprule
Platform & \shortstack{North\\RMSE (m)} & \shortstack{East\\RMSE (m)} & \shortstack{Down\\RMSE (m)} & \shortstack{Total\\RMSE (m)} \\
\midrule
PX4         & 0.145 & 0.098 & 0.414 & 0.450 \\
ROSflight   & 0.184 & 0.149 & 0.460 & 0.517 \\
Veloxity    & 0.179 & 0.161 & 0.075 & 0.252 \\
\bottomrule
\end{tabular}
\end{adjustbox}
\end{table}

Flight speed is configured for all three flights. Both reference flights used 3~m/s: configured through the ROScopter path manager for the ROSflight firmware and a mission speed command for PX4. The Veloxity flight used 4~m/s configured through the ROScopter path manager (Table~\ref{tab:outdoor_speed}). The higher limit demands faster attitude changes at each leg and leaves the inner loops less time to settle before the next waypoint. Despite Veloxity's more demanding setting, North and East errors are similar to those of both reference platforms, and a lower total RMSE is achieved thanks to lower vertical error. These results show that Veloxity brings Rust's memory-safety protections to the flight-control core with tracking performance comparable to the ROSflight firmware counterpart and to PX4. They also show Veloxity integrating in real time with ROSflight's existing autonomy interfaces. We note that the PX4 run's circuits are shorter because setpoints cut the corner at each waypoint instead of stopping and holding. 

\begin{table}[hbt!]
\caption{\label{tab:outdoor_speed} Mission speed settings and resulting flight speeds over the four circuits of Fig.~\ref{fig:outdoor_flights}. Speeds are computed from positions resampled every 0.2~s; flown speed uses the estimated position, reported at the 99th percentile to exclude estimator jitter.}
\centering
\small
\setlength{\tabcolsep}{4pt}
\begin{adjustbox}{max width=\textwidth}
\begin{tabular}{@{}lcccccc@{}}
\toprule
Platform & \shortstack{Speed \\setting (m/s)} & \shortstack{Peak commanded\\speed (m/s)} & \shortstack{Mean commanded\\speed (m/s)} & \shortstack{Flown speed,\\99th pct. (m/s)} & \shortstack{Mean circuit\\time (s)} & \shortstack{Four-circuit\\time (s)} \\
\midrule
PX4         & 3 & 3.04 & 2.11 & 3.06 & 31.5 & 126.0 \\
ROSflight   & 3 & 3.00 & 1.48 & 3.05 & 46.5 & 186.1 \\
Veloxity    & 4 & 4.00 & 1.99 & 3.95 & 34.5 & 138.0 \\
\bottomrule
\end{tabular}
\end{adjustbox}
\vspace{-20pt}
\end{table}

\section{Conclusion}

This paper presented Veloxity, an open-source UAV flight-control firmware implementation written in memory-safe Rust. Veloxity confines unsafe operations to explicitly marked regions so the compiler can enforce memory and type safety in the flight-control core. We evaluate it against the ROSflight firmware at multiple levels. In isolated numerical tests, its controller and mixer outputs effectively match the ROSflight firmware. In matched-noise software-in-the-loop simulation, its mean tracking error is statistically equivalent to ROSflight's within 1.3~cm (rounded). In outdoor flight on the same Pixracer Pro hardware, it sustains a 400~Hz firmware control rate, and tracks a reference waypoint mission with error comparable to both ROSflight and PX4 despite a higher speed setting. Throughout, Veloxity operated within the ROScopter autonomy stack. These results provide evidence that Rust is a practical alternative to C and C++ for resource-constrained, safety-critical flight firmware.

By providing a modular, extensible, and memory-safe foundation for developing and evaluating flight-control algorithms, Veloxity can support research into increasingly sophisticated autonomous systems and serve as a step toward safety-critical domains where software failure is unacceptable, such as civil aviation and commercial spaceflight. Because Veloxity integrates with ROSflight's existing autonomy, simulation, and hardware interfaces, researchers can adopt a memory-safe firmware without rewriting the rest of their stack. This lowered cost of moving toward memory-safe software in practice advances the rapid pace of safety focused research. Future work will extend Veloxity to additional flight controllers and fixed-wing vehicles and carry memory safety up through the companion computer into the remaining autonomy stack.

\section*{Code Availability}
Veloxity is publicly available under the
BSD 3-Clause License. Source code is available at \url{https://github.com/byu-magicc/Veloxity}, and documentation can be found at \url{https://magicc.byu.edu/Veloxity/}.

\section*{Acknowledgments}
OpenAI's ChatGPT and Anthropic's Claude models were used to assist in manuscript image generation, language refinement, and LaTeX formatting. Generative AI tools were also more broadly involved in the iterative, human-led development of Veloxity. All generated code and contributions were subject to human review, verification, and refinement, with the authors retaining responsibility for methodology, results, and conclusions.

\bibliography{references/references}

\end{document}